\documentclass[final,5p,times,twocolumn]{elsarticle}

\journal{Image and Vision Computing}

\usepackage{graphics} % for pdf, bitmapped graphics files
\usepackage{epsfig} % for postscript graphics files
\usepackage{mathptmx} % assumes new font selection scheme installed
\usepackage{times} % assumes new font selection scheme installed
\usepackage{amsmath} % assumes amsmath package installed
\usepackage{amssymb}  % assumes amsmath package installed
\usepackage{fixltx2e}
\usepackage{caption}
\usepackage{gensymb}
\usepackage{color}

\usepackage[hyphens]{url}
\usepackage{hyperref}
\hypersetup{breaklinks=true}

\begin{document}

\begin{frontmatter}

\title{Computer Vision in Automated Parking Systems: Design, Implementation and Challenges}

\author[mymainaddress]{Markus Heimberger}
\author[mysecondaryaddress]{Jonathan Horgan}
\author[mysecondaryaddress]{Ciaran Hughes}
\author[mysecondaryaddress]{John McDonald}
\author[mysecondaryaddress]{Senthil Yogamani\corref{mycorrespondingauthor}}
\cortext[mycorrespondingauthor]{Authors are listed in alphabetical order. Corresponding author's contact is}
\ead{senthil.yogamani@valeo.com}

\address[mymainaddress]{Automated Parking Product Segment, Valeo Schalter Und Sensoren, Bietigheim, Germany}
\address[mysecondaryaddress]{Automated Parking Product Segment, Valeo Vision Systems, Tuam,  Ireland}

\begin{abstract}
Automated driving is an active area of research in both industry and academia. Automated Parking, which is automated driving in a restricted scenario of parking with low speed manoeuvring, is a key enabling product for fully autonomous driving systems. It is also an important milestone from the perspective of a higher end system built from the previous generation driver assistance systems comprising of collision warning, pedestrian detection, etc. In this paper, we discuss the design and implementation of an automated parking system from the perspective of computer vision algorithms. Designing a low-cost system with functional safety is challenging and leads to a large gap between the prototype and the end product, in order to handle all the corner cases. We demonstrate how camera systems are crucial for addressing a range of automated parking use cases and also, to add robustness to systems based on active distance measuring sensors, such as ultrasonics and radar. The key vision modules which realize the parking use cases are 3D reconstruction, parking slot marking recognition, freespace and vehicle/pedestrian detection. We detail the important parking use cases and demonstrate how to combine the vision modules to form a robust parking system. To the best of the authors' knowledge, this is the first detailed discussion of a systemic view of a commercial automated parking system.
\end{abstract}


\end{frontmatter}

%\linenumbers

\section{Introduction}
Cameras have become ubiquitous in cars, with a rear-view camera being the minimum and full surround view camera systems at the top-end. Automotive camera usage began with single viewing camera systems for the driver. However, both the number of cameras and the number of ADAS applications made possible with automotive cameras have increased rapidly in the last five years, mainly due to the fact that the processing power has increased during this time period to enable the high levels of real-time processing for computer vision functions. Some examples include applications such as back-over protection, lane departure warning, front-collision warning, or stereo cameras for more complete depth estimation of the environment ahead of the vehicle. The next level of advanced systems require driving automation in certain scenarios like highway or parking situations. There are many levels of autonomous \footnote{The words autonomous and automated are used interchangeably by researchers in both industry and academia. In this paper, we use the term automated instead of autonomous implying that the system is not completely independent and there is a driver trigger.} driving as defined by Society of Automotive Engineers \cite{LevelsAD}. Fully autonomous driving (Level 5) is an ambitious goal. The current systems are, at best, Level 3 and the commercial deployment is mainly for highway driving. In this paper, we focus on Level 2 or Level 3 type automated parking systems. 

Certainly there are risks involved as no algorithm is perfect, and the sensors utilized can have limitations in certain scenarios. Automated parking is a good commercial starting point to deploy automated driving in a more restricted environment. Firstly, it involves low speed manoeuvring with a low risk of high impact accidents. Secondly, it is a more controlled environment with fewer scene variations and corner cases. Stable deployment of automated parking in the real world and analysis of performance statistics is an important step towards going to higher levels of autonomy.

\begin{figure}[!t]
\centering
\includegraphics[width=\columnwidth]{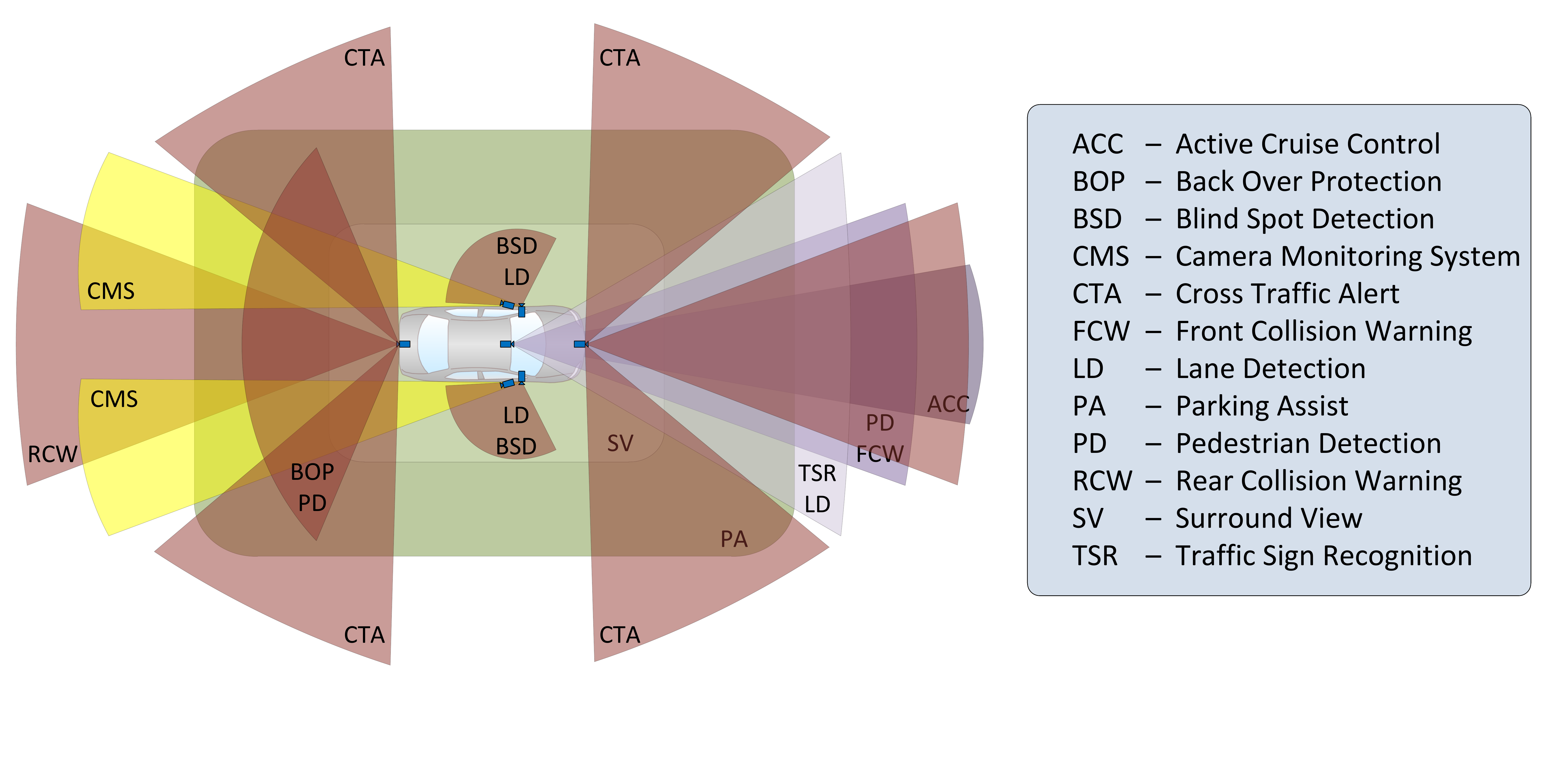}
\caption{Camera based ADAS applications and their respective field of view}
\label{fig:cameraFOV}
\end{figure}

\begin{figure}[!t]
\centering
\includegraphics[width=\columnwidth]{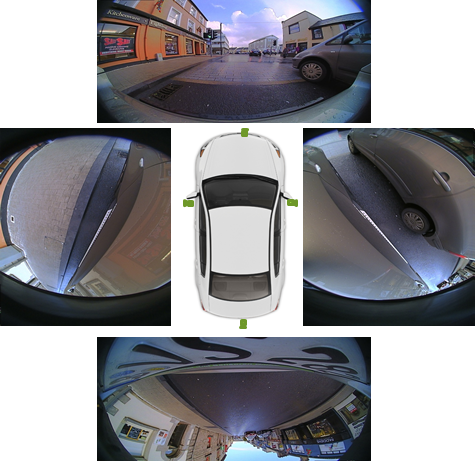}
\caption{Sample images from the surround view camera network demonstrating near field sensing and wide field of view}
\label{fig:svs}
\end{figure}

The first generation parking systems were semi-automated using ultrasonics or radar. Cameras are recently augmenting them to provide a more robust and versatile solution. In this paper, we consider cameras as an important component of a parking system, extending the capabilities of or providing inexpensive alternatives to other sensors. Figure \ref{fig:cameraFOV} shows the various field of views of common ADAS applications \cite{horgan2015vision}, some of which is needed for parking systems. Typically, surround view camera systems consist of four sensors forming a network with small overlap regions, sufficient to cover the near field area around the car. Figure \ref{fig:svs} shows the four views of a typical camera network such as this. It is important to note that the cameras are designed and positioned on the vehicle to maximise performance in near field sensing (which is important for automated parking). As part of this near field sensing design, they use wide-angle lenses to cover a large field of view (easily exceeding 180\degree horizontally). Thus, algorithm design must contend with fisheye distortion, which is not an insignificant challenge as most of the academic literature in computer vision is focused on rectilinear cameras or, at most, cameras with only slight radial distortion. 

Designing a Parking system has a multitude of challenges. There are high accuracy requirements because of functional safety aspects, risk of accident and consumer comfort (for example, the car cannot park such that a driver cannot open their door). The infrastructure is relatively unknown with possibility of dynamic interacting objects like vehicles, pedestrians, animals, etc. Varying environmental conditions could play a massive role as well. For instance, low light conditions and adverse weather like rain, fog can inhibit the accuracy and detection range significantly. There is also the commercial aspect that can bound the computational power available on a low power embedded system. On the other hand, the parking scenario is much more restricted in terms of the set of possibilities compared to full autonomous driving. Vehicle speeds are low, giving enough processing time for decisions. The camera motion is restricted with well-defined region of interest. There is possible assistance from infrastructure to ease this problem, especially to find and navigate to a empty parking slot \cite{mahmud2013survey}. While in this work, we don’t discuss any infrastructure support, the authors feel that this will be an important part of the automated parking solution.

The term automated parking can refer to a smart infrastructure which manage the placement of cars in a mechanical parking lot, typically multi-tiered or a smart electronic system embedded in a car. A simple literature search shows that majority of the results correspond to this meaning and not the meaning we use. \cite{wang2014automatic} and \cite{xu2000vision} are the closest to a full vision based automated parking system. These papers focus only on the computer vision algorithm. In contrast, in this paper, we aim to provide a more complete review of the use of computer vision in parking, in terms of detailing the use cases and expanding upon basic computer vision modules needed.

\subsection{Structure of this paper}

\begin{figure}[!t]
\centering
\includegraphics[width=\columnwidth]{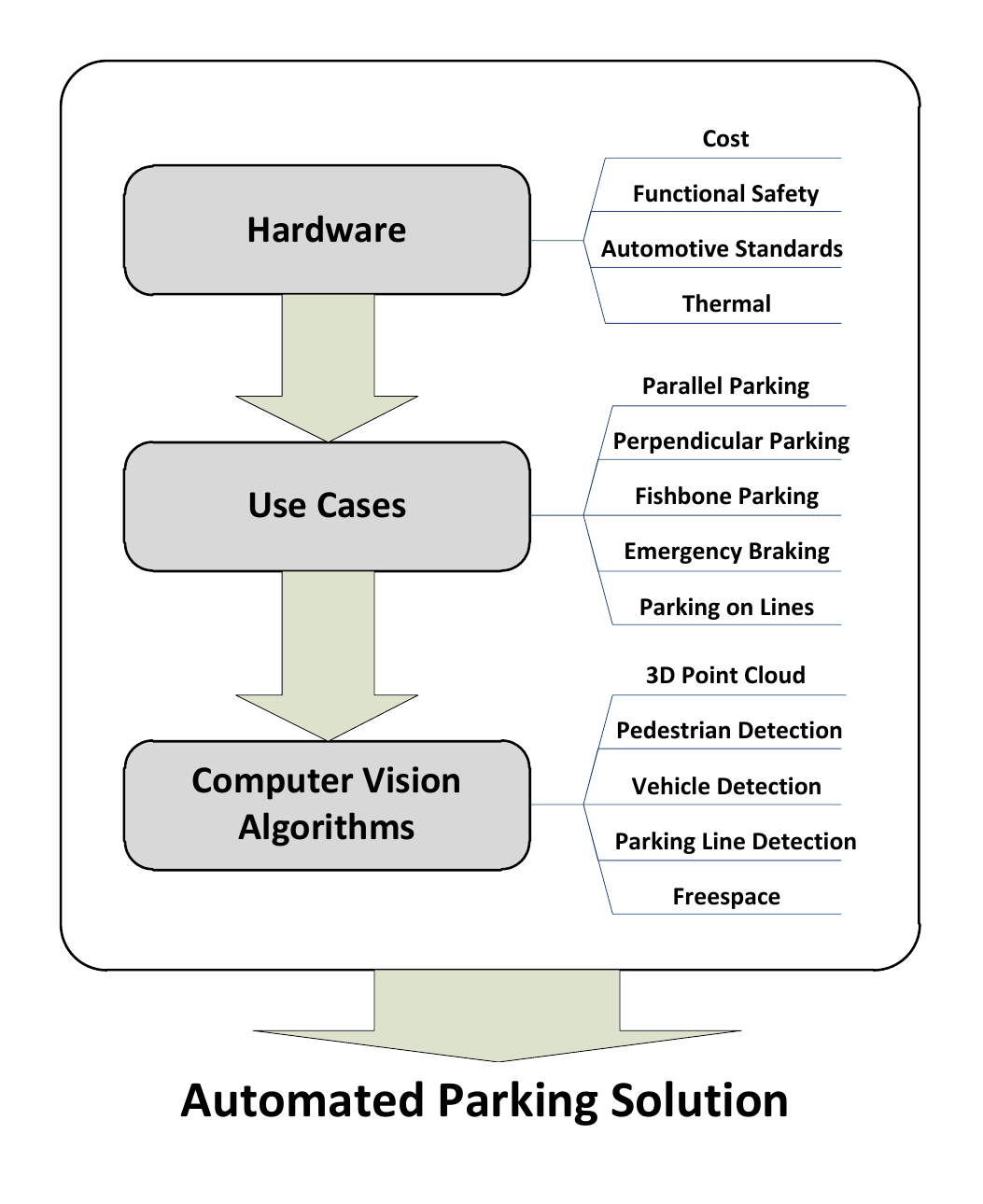}
\caption{Decision flow for design of camera parking system}
\label{fig:decisionFlow}
\end{figure}

Figure \ref{fig:decisionFlow} gives a high level overview of the decision flow when designing automated parking systems (and indeed, with adaptation, most ADAS functions), with some of the design decisions that need to be considered at each stage. The biggest limiting factor in design is the hardware choice, as automotive systems have harder constraints (such as cost, safety factors, standards adherence, thermal concerns and many others) than commercial electronic systems. For these reasons, we treat hardware first in Section 2, where we consider practical system considerations of ECU, cameras and processing components. Given the defined hardware limitations, the next step is to understand the use cases; i.e., what is the goal of the system, in terms of the end user functionality? Thus, Section 3 details the various important parking use cases and how each scenario can be handled by a vision system. Finally, with hardware limitations known and end user goals defined, the designer must select the appropriate algorithms to implement to achieve the system requirements. Section 4 discusses the various building block vision algorithms needed to realize a high level automated parking system. In Section 5, we return to system level topics, discussing how it all fits together, the various challenges with the limitations and take a glimpse into the next generation of vision functions for parking.

\section{Hardware components}

In this section, we provide an overview of the system components which make up an parking system. We highlight the role of safety aspects and computational limitations due to commercial aspects.

\subsection{ECU System and Interfacing electronics}

At high level, there are two type of camera system. Standalone camera with a small embedded system tightly integrated in the camera housing. This is sufficient for smaller applications like a Rear View camera. But for more complex applications, the camera is typically connected to a powerful external SOC via additional interfacing electronics. As illustrated in the Figure \ref{fig:svs}, for a typical surround view system with 4 camera inputs, the spatially separated cameras have to be connected to a central ECU. The data bandwidth requirements for video is high compared to other systems, which brings in a lot of challenges and limitations in SOC. The raw digital output from a sensor is typically 10/12-bit but the video input port of the SOC might support just 8-bit. This mandates an external ISP to compress the depth to 8-bit. Other simple factors like resolution and frame-rate could double the system requirements. The connectivity between SOC and camera is typically wired via twisted pair or coaxial cable. 

\begin{figure*}[!t]
\centering
\includegraphics[width=2\columnwidth]{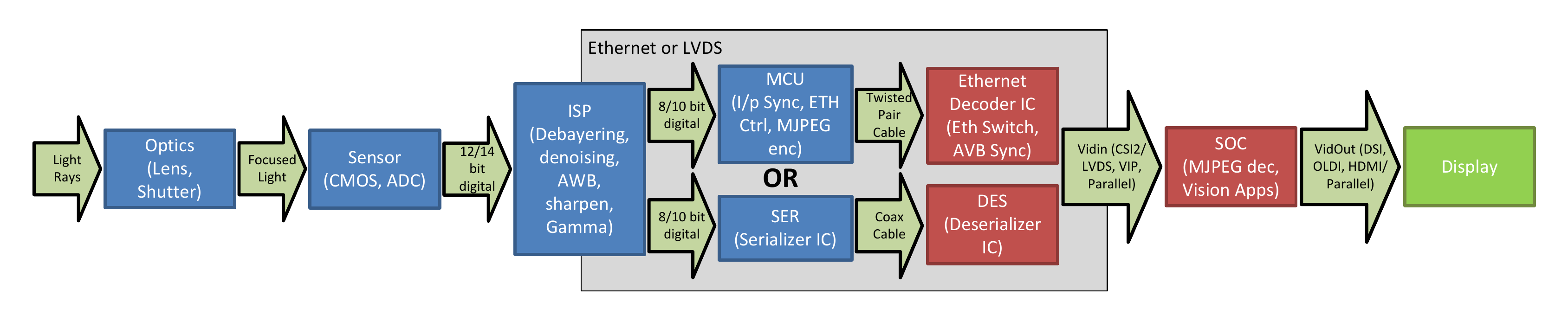}
\caption{Block diagram of a Vision System}
\label{fig:cameraISP}
\end{figure*}

Figure \ref{fig:cameraISP} illustrates the two alternate methodologies used. Use of serializer and deserializer (together known as SerDes) and signaling via co-axial cable is more common because of its high bandwidth of 1 Gbps/lane. Coaxial cable interfaces employ Fakra connectors as it is commonly used by European OEMs. Ethernet interface and twisted pair cable is a cheaper alternative but it has has a relatively limited bandwidth of 100 Mbps. To compensate for it, Motion JPEG is performed before transmission which causes a limitation of having the complete ISP separately and a conversion chip for MJPEG. The other alternative can leverage the SOC ISP. Ethernet cameras also require more complex electronic circuitry on both ends. Gigabit Ethernet could be used to achieve higher bandwidth but it is more expensive and defeats the purpose of lower cost. 

Most of the modern SOC interfaces are digital and serial.  MIPI (Mobile Industry Processor Interface) standardized the serial interfaces for camera input CSI (Camera serial interface) and DSI (Display serial interface). These interfaces are implemented as LVDS (Low-Voltage Differential Signalling) connectors underneath. CSI2 is the current generation with a bandwidth of 1 Gbps/lane.  OLDI (Open LVDS Display Interface) is the open LVDS interface which works on bare-metal LVDS.  Some SOCs provide parallel interfaces in addition to the serial interfaces. Although parallel interfaces provide a larger bandwidth, they require larger wiring and more complex circuitry which is not scalable.
     
Vehicle interfaces, such as CAN (Controller Area Network) or FlexRay, carry the signals from the car to the SOC. With respect to ADAS systems, odometry related signals like wheel speed, yaw rate, etc. are useful for algorithms requiring some knowledge of odometry. It could also provide signals like ambient light levels, fog/rain sensor, etc. which could be helpful to adapt the algorithm according to external conditions. The common communication protocols used are CAN and Flexray as they are low payload data. For, high payload signals, sometimes Ethernet protocol is used. Flexray is an improved version of CAN (faster and more robust) and hence more expensive as well. CAN FD (flexible data-rate) is an improved second generation of CAN. Many of the automotive SOCs have direct interface to CAN, while some additionally support FlexRay.

As mentioned before, memory is a critical factor in vision systems. There are several types of memory involved, the main memory is usually DDR (Double Data Rate) which typically starts at 256MB and could go to several GBs. The image and the intermediate processing data resides here. The high end current generation systems use DDR3 and will eventually move towards DDR4. There is also Flash/EEPROM memory for storing persistent data, like bootup code, configuration parameters and sometimes statistics of the algorithm outputs. On the SOC, there is on-chip memory (L3) in the order of few MBs which is shared across the different cores on the chip, which can be used as a high speed buffer to stream from DDR. There is also cache or internal memory inside the processors (L1 and L2) which has access rates close to the clock frequency of the processor. DMAs (Direct Memory Access) are common in vision systems for ping-pong buffering of data from DDR to L2/L3 memories.  It is important to appreciate the hierarchies of memory, which have opposite gradation in size and speed. They are arbitrated via a memory interface MEMIF in the SOC. Memory often becomes a serious bottleneck in such systems, a fact that is often not understood or overlooked. A detailed bandwidth analysis of the algorithms is necessary to decide the speed of memories and the bandwidth of MEMIF.

Debugging is typically done via JTAG and an IDE and this is usually not supported in the native ECU and a breakout board is necessary during development stage. For Ethernet systems, there is direct exposure of ECU memory via file systems. Sometimes debugging is also done through logging via UART. The other peripherals like SPI (for serial comm), I2C (master-slave electronics sync), GPIO (general purpose pins),etc are standard as in other electronic systems.

\begin{figure}[!t]
\centering
\includegraphics[width=\columnwidth]{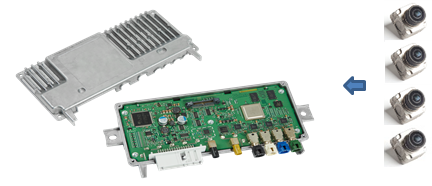}
\caption{ECU and camera housing}
\label{fig:ecu}
\end{figure}

\subsection{Camera}

The camera package typically consists of the imaging sensor, optical system and an optional ISP HW.

\textbf{Optics}: The optical system consists of lens, aperture and shutter. These components are captured in the camera matrix via focal length (f), aperture diameter (d), field of view (FOV) and Optical transfer function (OTF).

Role of MTF: Modulation Transfer Function (MTF) corresponds to the number of pixels that are exposed for capture by the camera. Many cameras have the option to select a subset of the available active pixels via setting appropriate registers on the image sensor. A higher number of active pixels can directly mean an improvement for computer vision algorithms in terms of range and accuracy of detection. However, it is important to remember that the resolution of the captured image is limited by both the number of available pixels and the optical resolution of the lens (as determined by the overall lens quality of the attached camera lens which is limited by diffraction of the elements in the lens). Additionally, the spatial resolution of a camera is impacted by increasing physical pixel size of the sensor \cite{farrell2006resolution}. The overall resolution of the camera and lens combination can be measured using by MTF \cite{boreman2001modulation}. 

Fisheye lenses: Fisheye lenses are commonly used in automotive to obtain a larger FOV. This produces non-linear distortion in the images which is typically corrected for viewing functions. For the processing part, due to the noise incurred by un-distortion of low resolution areas (towards the periphery) to higher resolution in linear image, sometimes it is more suitable to run the algorithm directly on the fisheye image. Typical, forward ADAS functions, such as front collision warning, lane departure warning, and head light detection, will use lenses with narrow fields of view (such as 40\degree to 60\degree). However, short range viewing, such as top-view and rear-view, and detection application, such as back over protection and pedestrian detection, require cameras that provide a much wider field of view (Figure \ref{fig:svs}). The use of wide-angle lenses, however, introduces complications in lens design which leads to the mathematics that describe the camera projections being significantly more complex. Basically, a straight line in the world is no longer imaged as a straight line by the camera – geometric distortion is introduced to the image. A detailed overview of the application wide-angle lenses in the automotive environment is given in \cite{hughes2009wide}.

\textbf{Sensor}: Omnivision and Aptina are the commonly used sensor vendors, though other manufacturers are available. Visual quality of cameras has been improving significantly. The main factors that influence the systems design from the camera selection are resolution (1 MP to 2 MP, and higher), frame rate (30 to 60 fps) and bit depth (8 to 12 bit). There is clear benefit in improving these, but they come with significant overheads of memory bandwidth.

Dynamic Range: Dynamic range of an image sensor describes the ratio between the lower and upper limits of the luminance range that the sensor can capture. Parts of the scene that are captured by the image sensor below the lower limit will be clipped to black or will be below the noise floor of the sensor, and conversely, those parts above the higher limit will be saturated to white by the image sensor. There is no specific threshold for dynamic range at which a sensor become High Dynamic Range (HDR), rather the term is usually applied to an image sensor type that employs a specific mechanism to achieve a higher dynamic range than conventional sensors. Note that the upper and lower luminance limits for an image sensor is not fixed. Indeed, many sensors can dynamically adapt the limits by altering the exposure time (also called shutter speed) of the pixels based on the content of the scene – a bright scene will typically have a short exposure time, and a dark scene will have a long exposure time. As a basic ratio, the dynamic range of a sensor is typically given in dB. Dynamic range is important in automotive vision, as due to the unconstrained nature of automotive scenes, often there will be a scenario with high dynamic range. Obvious examples of high dynamic range scenes are when the host vehicle is entering or exiting a tunnel, or during dusk and dawn when the sun is low in the sky.

Sensitivity: The sensitivity of a pixel measures the response of the pixel to illuminance over a unit period of time. Many things can impact the sensitivity of a pixel, such as silicon purity, pixel architecture design, microlens design, etc. However, one of the biggest factors is simply the physical size of the pixel. A pixel with a higher area will have the ability to gather more photons, and thus will have a greater response to lower illuminance. However, increasing sensitivity by increasing pixel size will have the impact of reducing the spatial resolution \cite{farrell2006resolution}.

Signal to Noise Ratio: Signal to noise ratio is probably the most intuitive property for an engineer coming from a signal processing background. It is the ratio of the strength (or level) of the signal compared to sources of noise in the imager. The primary issue is that the methods that image sensor manufacturers use to measure noise is non-standard, so drawing comparisons between different image sensor types based on SNR is difficult. Additionally, the SNR advertised will be based on a fixed scene, whereas the actual SNR of the image received will be scene dependent, and influenced by the pixel exposure time and gain factors applied to the signal, among other factors. For example, a dark scene in which the camera has longer exposure time and higher gain factors applied to the output will result in an image with a low SNR. Temperature typically plays a major role in the level of noise in an image and thermal management of a camera device plays a critical role in reducing the  amount of noise present in an output image. This can be aided by designing a camera system that keeps the image sensor isolated as much as possible from sources of heat. The level of noise in an image has a deteriorating effect on the performance of vision algorithms.

Frame Rate: The maximum frame rate of a sensor has a direct influence on the time response of an algorithm. For algorithms designed to work at higher vehicle velocities, higher frame rate is important as the necessary response times of the system will be lower. However, there are system considerations that need to be taken into account. For example, a higher frame rate means that you will have a shorter time period in which to apply any algorithms to the image for your application. But this will require a more powerful and expensive image processing hardware. Additionally, an increased frame rate will result in a lower maximum exposure time for the image sensor pixels (33.3ms at 30 fps versus 16.6ms at 60fps), which will have a direct impact on the performance of the image sensor in low light scenarios.

\textbf{ISP}: Converting the raw signal from sensor to viewable format includes various steps like debayering, denoising and High Dynamic Range processing. These steps are collectively referred to as Image Signal Processing (ISP). Most of the ISP is typically done in HW either in the sensor itself, as a companion chip ISP or in the main SOC (System on Chip). ISP is, fundamentally, the steps that are required to convert the captured image to its usable format by the application. For example, most colour image sensors employ the Bayer colour filter, in which alternate pixels in the imager have a red, green or blue filter to capture the corresponding light colour. To get a usable/viewable image (e.g. full RGB or YUV), debayering is necessary. Other typical ISP steps include, for example, denoising, edge enhancement, gamma control and white balancing. Additionally, HDR image sensors will need a method to combine two or more images of different exposure time to a single, HDR image. Most ISP is typically done in HW, either in the sensor chip itself as an SOC (such as the OV10635 or MT9V128), within a companion chip ISP (such as the OmniVision OV490 or the Aptina AP0101AT), or in SOC with the main processing unit (such as the Nvidia Tegra). Of course, additional or custom image post-processing can be done in a generic/reconfigurable  processor, such as GPU, DSP or FPGA. The level of required ISP is completely application dependent. For example, many typical ADAS applications require only grey scale images, in which case a sensor without the Bayer filter array could be employed, which would subsequently not require the debayering ISP step. Additionally, several of the ISP steps are designed to provide visual brilliance to the end user for viewing applications. This may be unnecessary or even counter-productive for ADAS applications. For example, edge enhancement is employed to give sharper, more defined edges when the image is viewed, but can have the result that edge detection and feature extraction in ADAS applications is less accurate.

\subsection{SOC}

The following discussion is based on our experience working with various SOCs (System on Chip) targeted for ADAS high-end market. This is by no means an exhaustive comparison; we have left out big SOC players like Intel and Qualcomm who are not popular in ADAS and the generations could be different. We have summarized the different types of processing units which are relevant for vision algorithms below.

\textbf{GPP} (General Purpose Processor) is typically the master control processor of an SOC. Some flavour of ARM Cortex Application processors is commonly used as the GPP. The flavours vary from A9, A15 to A53 and A57. The former two are the popular GPP in the current generation devices and the latter two are 64-bit roadmap processors. NEON is a SIMD engine which can accelerate image processing. Because of the ubiquitous nature of multimedia, importance of NEON has grown and it is becoming more tightly integrated. The only major exception of an SOC not using ARM is Mobileye EyeQ which uses MIPS cores instead. Toshiba’s TMPV7600 uses their proprietary MeP low power RISC cores but they use A9 in some of the TMPV75 processors.

\textbf{GPU} (Graphics Processing Unit) was traditionally designed for graphics acceleration. It can be used for view rendering/remapping and for drawing output overlays using OpenGL. Some of the GPUs can perform only 2D graphics via OpenVG. They are slowly being re-targeted to be used as additional resource for vision algorithms through OpenCL. Nvidia is leading this way providing a General Purpose GPU processor for vision. With respect to Vision algorithms, CUDA (Compute Unified Device Architecture) of Nvidia is significantly more powerful than other GPUs provided by ARM (Mali), PowerVR (SGX) and Vivante. The performance power of Nvidia GPUs are growing at a significantly faster rate compared to other processors to suit the growth in the field of automotive vision applications. CUDA uses SIMT (single instruction multiple threads) with threading done in hardware. It has a limitation of dealing with load/store intensive operations which is improved by high data-bandwidth provided to it.

\textbf{SIMD} engines are quite popular in the application areas of image processing. This is because the input data is 8-bit fixed point and the initial stages of image processing are typically embarrassingly parallel. These processors are typically designed to be power-efficient by avoiding floating-point and having a simplified processing pipeline. For instance, performance/watt of TI's Embedded Vision Engine (EVE) \cite{mandal2014embedded} is $\sim$8X than that of A15. On the other hand, it lacks flexibility and it is typically used for the pixel-level initial stages of the pipeline. TI and Renesas have small width (8/16) SIMD processors where Mobileye and Freescale make use of a large SIMD width (76/128) processor.

\textbf{DSPs} (Digital Signal Processors) are traditional processors designed for low-power acceleration of signal processing algorithms. Most of them including TI's C66x and Toshiba's MPE are VLIW (Very Large Instruction Word) based superscalar processors. VLIW is an efficient architecture scheme to provide flexibility compared to SIMD. They also exhibit a form of SIMD using packed arithmetic i.e. aliasing 4 byte instructions in a word instruction.  The gap of performance normalized to cost of DSP is becoming closer because of lowered cost and performance improvements of NEON and its ubiquity.

\textbf{ASIC} (Application-specific Integrated Circuit) implements the entire algorithm in hardware with minimal flexibility like modification of parameters coded via register settings.  If the algorithm is standardized, ASIC provides the best performance for the given silicon. But it requires a lot of initial investment which impacts time to market and there is a risk of being redundant particularly for the area of computer Vision where the algorithms are not standardized and could change drastically. Mobileye is one company who primarily uses ASIC. But moving from EyeQ3 to EyeQ4, they have shifted to flexible processors.  Toshiba’s TMP device has HW acceleration for object detection feature (HOG) and 3D reconstruction (SFM) and Renesas provides an IMR HW module to perform distortion correction.

\textbf{FPGA} (Field Programmable Gate Array) is closely related to ASICs but has a key advantage of being re-configurable at run-time. From the application perspective, this means the HW can adaptively transform to an accelerator needed for that particular scenario. For instance, if the algorithm has to be drastically different for low and high speeds, the FPGA can transform into one of these as required. But because ASIC is hard-coded to be optimized for one algorithm, FPGA will be lagging in performance. Due to recent progress in FPGA technology, the gap is closing.  Xilinx and Altera are the major FPGA suppliers who offer very similar performance FPGAs. 

\textbf{Others}: Many of the SOCs provide video codecs (H.264), JPEG and ISP as ASICs. The video codec provide Motion Estimation (Block-matching Optical Flow) which is very useful for the vision algorithms. Some SOCs provide a safety compliant micro-processor like Cortex-M4 or R4 for AUTOSAR (AUTomotive Open System ARchitecture) and improved ASIL.

\textbf{Summary}: Typical design constraints for SOC selection for embedded systems are performance (MIPS, utilisation, bandwidth), cost, power consumption, heat dissipation, high to low end scalability and programmability.  Unlike hand-held devices, power consumption is not a major criterion as it is powered by car battery. Heat dissipation matters only up to a threshold and might add to costs through better heat sinks. Programmability is becoming abstracted via SW frameworks like OpenCL and is not a major factor in cost. Hence for ADAS, the main factors finally boil to cost and performance. Because of the diverse nature of the processors, this is typically a difficult decision to make. Usually comparing the processor via MIPS is not useful as the utilization is heavily dependent on the nature of the algorithm. Hence a benchmark analysis for the given list of applications based on vendor libraries and estimates is critical for choosing an appropriate SOC. A hybrid architecture which combines fully programmable, semi programmable and hard-coded processors could be a good amortized risk option. Examples of commercial automotive grade SOCs are Texas Instruments TDA2x, Nvidia Tegra X1, Renesas R-car H3, etc.

\section{Automated Parking Use Cases}

\subsection{Overview of parking manoeuvres}

\begin{figure}[!t]
\centering
\includegraphics[width=0.8\columnwidth]{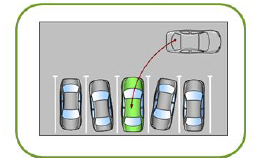}
\caption{Accurate vehicle parking based on slots, not other vehicles}
\label{fig:parking-accurate}
\end{figure}

Automated parking systems have been on the mass market for some time, starting with automated parallel parking and then advancing in more recent years to include perpendicular parking. Parking systems have evolved beyond driver assistance systems in which only steering is controlled (SAE Automation Level 1), to achieve partial automation of both lateral and longitudinal control \cite{sae2014taxonomy}. The challenge of parking assistant systems is to reliably and accurately detect parking slots to allow parking manoeuvres with a minimum amount of independent movements. The aim of an automated parking system is to deliver a robust, safe, comfortable and most importantly useful function to the driver enabling time saving, accurate and collision free parking. Current systems on the market rely solely on range sensor data, typically ultrasonics, for slot detection, remeasurement and collision avoidance during automated parking. While such systems have proven to be very successful in the field and are continuing to penetrate the market with releases in the mid to low end of the market they possess some inherent limitations that cannot be resolved without the help of other sensor technologies. The use cases described below focus on the benefits of camera based solutions with ultrasonic sensor fusion in particular to try and tackle some of the restrictions of current systems to move automated parking technology to the next step.

Before an automated parking manoeuvre can begin the parking system must first search, identify and accurately localise valid parking slots around the vehicle. Current systems typically rely on the driver to initiate the search mode. Parking slots can come in different forms as described in the use cases below. After the slots are located they are presented to the driver in order to allow selection of the desired parking slot as well as direction the car faces in the final parked position. After the driver has selected a parking slot the vehicle automatically traverses a calculated trajectory to the desired end location while driving within a limited speed range typically under 10kph. In order to maintain this function at automation level 2 to avoid the legal implications of the jump to conditional automation, where the system is expected to monitor the driving environment, the driver is required to demonstrate his attentiveness through the use of a dead man switch located in the vehicle \cite{sae2014taxonomy}. Partially automated systems can also allow the driver to exit the vehicle and initiate the parking manoeuvre remotely through a key fob or smart phone after the parking slot has been identified. In this case, the driver remains responsible for monitoring the vehicle's surroundings at all times and the parking manoeuvre is controlled through a dead man switch on a key fob or smart phone. Remote control parking is applicable in scenarios where the parking space has already been located and measured or in controlled environments, (such as garage parking), where the vehicle can be safely allowed to explore the environment in front with limited distance and steering angle.

  During the parking manoeuvre the system continues to remeasure the location of both the intended parking slot and the ego vehicle itself. Continued remeasurement during the manoeuvre is required to improve the accuracy of the end position due to slot measurement inaccuracy and ego odometry measurement error as well as to avoid any collisions with static or dynamic obstacles such as pedestrians.. The  parking trajectory is calculated in such a way that the most appropriate one to the parking situation is chosen, i.e. trajectory is selected to finish in the middle of the parking slot from the current position without any collision and a finite amount of manoeuvres/direction changes (i.e. drive to reverse and vice versa). An enhancement of the automatic parking system functionality is not only to park into the slot but also to park out from the slot.  

\subsection{Benefits of Cameras for Parking}

Current systems rely on range sensor information, typically ultrasonics sensors to identify and localise the parking slots. There are many inherent issues with range sensors for automated parking that can be partly or fully overcome with the use of camera data. Camera data in this case would ideally be from four surround view fisheye ($\sim$190\degree) cameras located in the the front and rear as well as both mirrors in order to aid both slot search, parking automation and also visualisation for all parking use cases. A single rear-view fisheye camera is also beneficial in a limited number of reverse parking use cases after slot localisation was already been performed by other sensors. Narrower field of view front cameras have little benefit to slot search but similar to rear view could help in the automation of forward parking use cases. The computer vision functions that aid in the these use cases are discussed in more detail in the next section.

\begin{figure}[!t]
\centering
\includegraphics[width=\columnwidth]{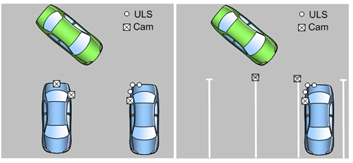}
\caption{Highlighting the benefit of using computer vision in fusion with traditional ultrasonics-based parking systems. (a) Increased detection performance and range, (b) detection of the environment not feasible with ultrasonics (lane markings)}
\label{fig:parking-cvbenefits}
\end{figure}

The biggest limitation of range sensors for slot detection is that they require other obstacles in the scene to identify the bounds of the parking slot. Cameras can be used to detect parking slots using the line markings on the road while taking advantage of the line ending type to understand the intended slot usage. It is possible to detect parking slot markings using LIDAR technology however, the sensor cost and the limited field of view are the big disadvantages. Figure \ref{fig:parking-accurate} illustrates that using a camera-fusion system the vehicle is parked with more accuracy. Based on ultrasonics/radar alone, the parking system would attempt to align with the other (inaccurately) parked vehicles while a camera/fusion allows parking against the slot itself. While the detection range of cameras ($\sim$10m) in terms of point cloud data for slot detection is less than that of radar or LIDAR ($\sim$100m) systems, cameras do provide a greater range than ultrasonics sensors ($\sim$4m)  while also having an overlapping field of view. Ultrasonics naturally provide accurate range data while cameras are more suited to providing high angular resolution, these attributes make the sensing capabilities of cameras and ultrasonics complementary. The extra detection range of cameras can provide the benefit of improved slot orientation accuracy as well as better slot validation particularly for perpendicular slots where orientation and range are more critical factors. Fisheye camera’s large vertical field of view ($\sim$140\degree) enables object detection and point cloud generation for obstacles above the height of the car within a close range (\textless 1m). This is beneficial for automated parking situations such as entering garages with roller doors where the door has not been opened sufficiently to allow the vehicle to enter. Most range sensors have a very restrictive vertical field of view and therefore cannot cover this use case.

Due to camera's significant measurement resolution advantage (1-2MP), they are capable of generating point cloud data for certain object types that active sensors may fail to detect such as poles or chain link fences. These “blind spots” for sensors such as ultrasonics can have a large impact on the robustness and reliability of the automated parking function. Surround view cameras can generate an accurate ground topology around the vehicle to aid in the localisation of kerbs, parking blocks and parking locks as well as surface changes for understanding of freespace. The large amount of data makes cameras very suitable for machine learning techniques allowing for object classification such as pedestrian and vehicles with an accuracy not equalled by any other sensor type. Classification of such objects generates another fusion data source resulting in a more intelligent and reactive automated parking system. Classification allows for intelligent vehicle reaction depending on object type, for example trajectory planning should not be made around a pedestrian as it is a dynamic object that can move in unrestricted and unpredictable way in the world. The accuracy of the vehicle's odometry information is vital for accurately detecting and localising the ego vehicle and also for smoothly traversing the parking trajectory with as few direction changes as possible. Cameras can be used to provide a robust source of vehicle odometry information through visual simultaneous localisation and mapping (SLAM) techniques made popular in robotics. This visual odometry can overcome many of the accuracy issues inherent in mechanical based odometry sensors and provide the resolution required to minimise parking manoeuvres updates after the initial slot selection. 

\subsection{Classification of parking scenarios}

\begin{figure}[!t]
\centering
\includegraphics[width=\columnwidth]{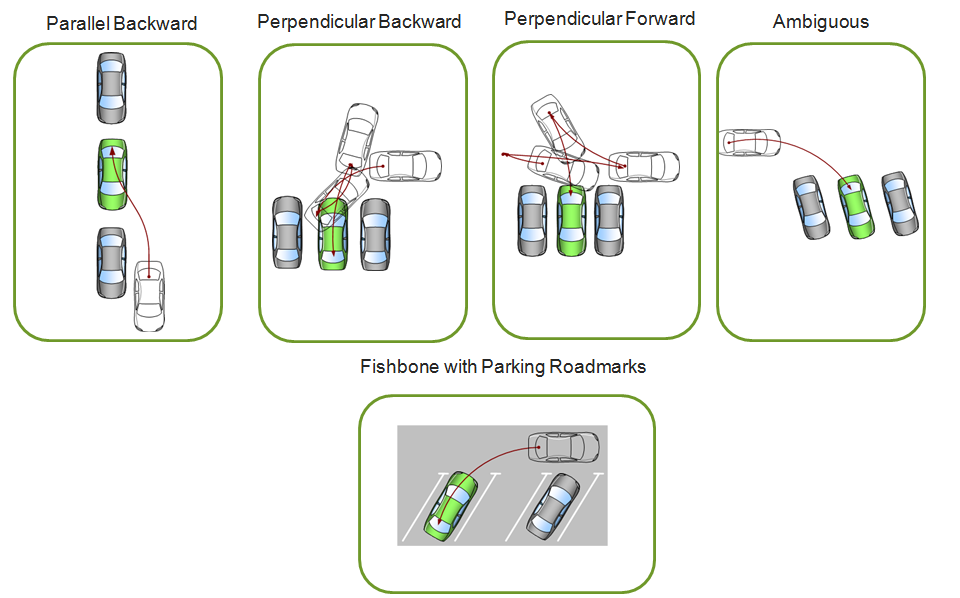}
\caption{Classification of Parking scenarios - (a) Parallel Backward Parking, (b) Perpendicular Backward Parking, (c) Perpendicular Forward Parking, (d) Ambiguous Parking and (e) Fishbone Parking with roadmarkings. }
\label{fig:parking-classification}
\end{figure}

There are various uses of autonomous parking, but in principle they can be classified in four main parking use cases:

  1. \textbf{Perpendicular Parking (forward and backward):} The system detects a parking slot laterally to vehicle as it passes by detecting the objects locations and line markings in the near field and measuring the slot size and orientation to understand if it can be offered to the user. If selected by the user for parking the systems finds a safe driving trajectory to the goal position of the parking slot while either orienting the vehicle relative to the slot bounds created by the other objects (vehicles in this case) or line markings. Figure \ref{fig:parking-classification} (b) describes an example of a backward parking manoeuvre completed in three steps, and the second part (c) describes a forward parking manoeuvre. Computer vision methods supports the detection of obstacles through both classification and SFM techniques. This data enhances the system detection rate and range in fusion with traditional ultrasonics-based systems (Figure \ref{fig:parking-cvbenefits} (a)), allowing for increased true positives and reduced false positives on slot offerings to the user while also improving slot orientation and measurement resulting in reduced parking manoeuvres. Computer vision also enables the parking of the vehicle based on parking slot markings, giving more accurate parking results (Figure \ref{fig:parking-cvbenefits} (b)), which isn’t feasible in traditional parking systems based on ultrasonics.

2. \textbf{Parallel parking:} Parallel parking (Figure \ref{fig:parking-classification} (a)) , like perpendicular parking, is a well defined parking situation. However, the manoeuvre and situation is significantly different. Typically, entering the parking space is completed in a single manoeuvre, with further manoeuvres used to align to the parking space more accurately. Additionally, parking tolerances are typically lower because of the desire to park close to the surrounding vehicles and the kerb inside the parking slot. Fusion with camera systems allows lower tolerances on the parking manoeuvres, and more reliable kerb detection (detecting kerbs is possible with ultrasonics and radar, but is often unreliable).

3. \textbf{Fishbone parking:}  Figure \ref{fig:parking-classification} (e) shows an example of fishbone parking where current ultrasonics based parking systems are limited, as the density of detections is too low to identify the orientation of the parking slot. In this case, using camera systems enables increased range to view inside the slot to determine the parking slot target orientation from both the objects or the line markings. This use case cannot be covered by current ultrasonics based systems.

4. \textbf{Ambiguous parking:} The final broad category of use case is the ambiguous parking situation, i.e. where the parking space is not well defined except by the presence of other vehicles and objects (Figure \ref{fig:parking-classification} d) ). The use of cameras enables advanced planning of the parking manoeuvre due to the increased detection range and more complete sensor coverage around the vehicle (ultrasonics typically do not cover vehicle flank) and thus enables more appropriate vehicle reaction in somewhat ill-defined use cases.

Additionally, the use of camera systems in parking enables or improves reliability of other functions, in comparison to ultrasonics/radar only parking systems, for example:

1. Emergency Braking/ Comfort Braking: Of course, with any level of autonomy, the vehicle needs to react to the presence of vulnerable road users. Sometimes, the environment can change quickly (e.g. a pedestrian quickly enters the area of the automatically parking vehicle), and as such, the vehicle must respond quickly and safely. By complementing existing parking systems, low speed automatic emergency braking or comfort braking is made significantly more robust due to the natural redundancy provided by camera fusion.

2. Overlaying of the object distance information: A very common use to combine vision system data with traditional parking systems is to overlay the object distance information in the video output stream e.g. in a surround view system. This helps the driver during manual vehicle manoeuvring to correctly estimate distance in the 360\degree video output stream for more precise navigation within the parking slot. This is especially helpful in parallel parking slot with a kerb, where the kerb is not visible to the driver. 

\section{Vision Applications}

Vision based ADAS applications first started appearing in mass production in early 2000s with the release of systems such as lane departure warning (LDW) \cite{sivaraman2013integrated}. Since then, there has been rapid development in the area of vision based ADAS. This is due to the vast improvements in processing and imaging hardware, and the drive in the automotive industry to add more ADAS features in order to enhance safety and improve brand awareness in the market. As cameras are rapidly being accepted as standard equipment for improved driver visibility (surround view systems), it is logical that these sensors are employed for ADAS applications in parallel. In this section, we will discuss the use of four important ADAS functions and their relevance in the context of automated parking systems. The focus is on algorithms that are feasible on current ADAS systems, considering the limitations described in Section 2, leaving treatment of state of the art algorithms (such as Deep Learning) to discussion in Section 5. Considering the functionality described in Section 3, we need to consider the detection, localisation and in some cases classification of 1) unmoving obstacles, such as parked vehicles, 2) parking lines and other ground markings, 3) pedestrians and generic moving obstacles, and 4) freespace to support the removal of tracked obstacles from parking map, for example. The algorithms discussed in the following sections are based on feasibility of deployment on embedded systems available two years ago. As the parking system has safety restrictions, going from a demonstration to a product takes a long cycle of iterative validation and tuning to get robust accuracy. We briefly discuss the current state-of-the-art in section \ref{sec:future}. 

\subsection{3D point cloud}

Depth estimation refers to the set of algorithms aimed at obtaining a representation of the spatial structure of the environment within the sensor’s FOV. In the context of automated parking, it is the primary mechanism by which computer vision can be used to build a map. This is important for all parking use cases: it enables better estimation of the depth of parking spaces over the existing ultrasonic-based parking systems, and thus better trajectory planning for both forward and backward perpendicular and fishbone park manoeuvring; it increases the reliability of kerb detection, improving the parallel parking manoeuvre; and, it provides an additional detection of obstacles, which, in fusion, reduces significantly the number of false positives in auto emergency braking.

\begin{figure}[!t]
\centering
\includegraphics[width=\columnwidth]{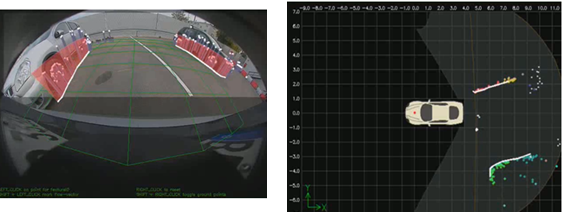}
\caption{Reprojected and top view of 3D reconstruction}
\label{fig:3dod}
\end{figure}

Depth estimation is the primary focus of many active sensor systems, such as TOF (Time of Flight) cameras, lidar and radar, this remains a complex topic for passive sensors such as cameras. There are two main types of depth perception techniques for cameras: namely stereo and monocular \cite{hartley2003multiple}. The primary advantage of stereo cameras over monocular systems is improved ability to sense depth. It works by solving the correspondence problem for each pixel, allowing for mapping of pixel locations from the left camera image to the right camera image. The map showing these distances between pixels is called a disparity map, and these distances are proportional to the physical distance of the corresponding world point from the camera. Using the known camera calibrations and baseline, the rays forming the pixel pairs between both cameras can be projected and triangulated to solve for a 3D position in the world for each pixel. Figure \ref{fig:3dod} shows an example of sparse 3d reconstruction.

Monocular systems are also able to sense depth, but motion of the camera is required to create the baseline for reconstruction of the scene. This method of scene reconstruction is referred to as structure from motion (SFM). Pixels in the image are tracked or matched from one frame to the next using either sparse or dense techniques. The known motion of the camera between the processed frames as well as the camera calibration, are used to project and triangulate the world positions of the point correspondences. Bundle adjustment is a commonly used approach to simultaneously refine the 3D positions estimated in the scene and the relative motion of the camera, according to an optimality criterion, involving the corresponding image projections of all points.

\subsection{Parking Slot Marking Recognition}

The detection of parking slots is, of course, critical for any automated parking system - the system must know where it will park ahead of completing the manoeuvre. To enable the detection of parking slots in the absence of obstacles defining the slot, and to enable more accurate parking, the detection of the road markings that define parking slots is critical. Consider this: in an empty parking lot, how would an automated parking system be able to select a valid parking slot? This is applicable for all parking manoeuvres (perpendicular, parallel and fishbone) in which parking against markings is required.

\begin{figure}[!t]
\centering
\includegraphics[width=\columnwidth]{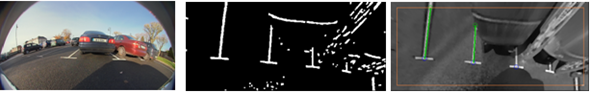}
\caption{Example of parking slot marking recognition}
\label{fig:psmd}
\end{figure}

It is possible to complete parking slot marking recognition using technology such as lidar, which has a spectral response on the road markings, as exemplified in \cite{hata2014road}. However, lidar systems are typically expensive, and suffer from limited detection areas, typically have a very narrow vertical field of view \cite{VelodyneDatasheet} compared with what is feasible with cameras (\textgreater 140\degree FOV). In vision, lane marking detection can be achieved using image top-view rectification, edge extraction and Hough space analysis to detect markings and marking pairs \cite{jung2006parking}. Figure \ref{fig:psmd} gives an example of the results from a similar approach, captured using 190\degree horizontal field of view parking camera \cite{liu2008bird}. The same authors also propose a different approach based on the input of a manually determined seed point, subsequently applying structural analysis techniques to extract the parking slot \cite{jung2006structure}. Alternatively, a pre-trained model-based method based on HOG (Histogram of Oriented Gradients) and LBP (Local Binary Patterns) features, with linear SVM (Support Vector Machine) applied to construct the classification models is proposed in \cite{su2014system}.

Regardless of the specific approach taken, what is clear is that the detection of marking slots is critical for a complete automated parking system, and that the only reasonably valid technology to complete this is parking cameras with a wide field of view.

\subsection{Vehicle and Pedestrian Detection/Tracking}

\begin{figure}[!t]
\centering
\includegraphics[width=\columnwidth]{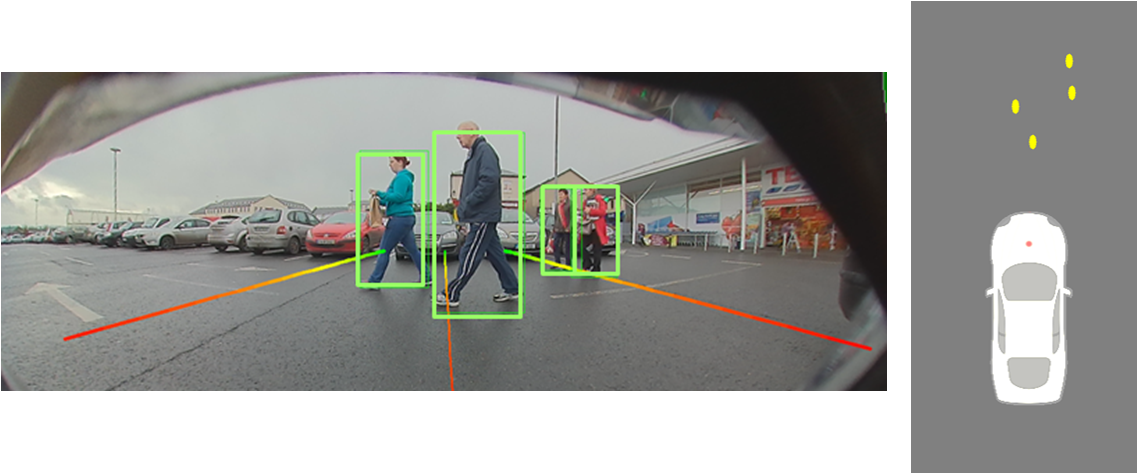}
\caption{Pedestrian classification and tracking using a parking camera}
\label{fig:pd}
\end{figure}

Vehicle detection and tracking is often done in the context of front camera detection \cite{sivaraman2013looking} for applications like auto emergency braking or in traffic surveillance applications \cite{buch2011review}.  However, parking manoeuvres are often done in the presence of other vehicles, either parked or moving, and as such, the detection and tracking of vehicles is important for the automation of such manoeuvres \cite{broggi2014vehicle}. Perhaps of higher importance in parking manoeuvres is for the system to reliably be able to both detect and classify pedestrians \cite{dollar2012pedestrian}, such that the vehicle can take appropriate action, e.g. auto emergency braking in the presence of pedestrians that are at potential risk \cite{dollar2012pedestrian} (Figure \ref{fig:pd}). Typically, both the problem of vehicle detection and that of pedestrian detection are solved using some flavour of classification. No other sensor can as readily and reliably classify detection based on object type, compared to vision systems.

Object classification generally falls under supervised classification group of machine learning algorithms. This is based on the knowledge that a human can select sample thumbnails in multiple images that are representative of a specific class of object. With the use of feature extraction methods such as histogram of oriented gradients (HOG), local binary patterns (LBP), and wavelets applied to the human classified sample images, a predictor model is built using machine learning in order to classify objects. Many vision based ADAS functions use machine learning approaches for classification. As already discussed, classification is extensively used in pedestrian and vehicle detection, but also in areas such as face detection and traffic sign recognition (TSR). The quality of the final algorithm is highly dependent on the amount and quality of the sample data used for learning the classifiers, as well as the overall quality of the classification technique and the appropriateness of the feature selection method for the target application. Typical classifiers include SVMs, random forest and convolutional neural networks (CNN). Recently there has been a shift in this trend via deep learning methods wherein the features are automatically learned.

\subsection{Freespace}

\begin{figure}[!t]
\centering
\includegraphics[width=\columnwidth]{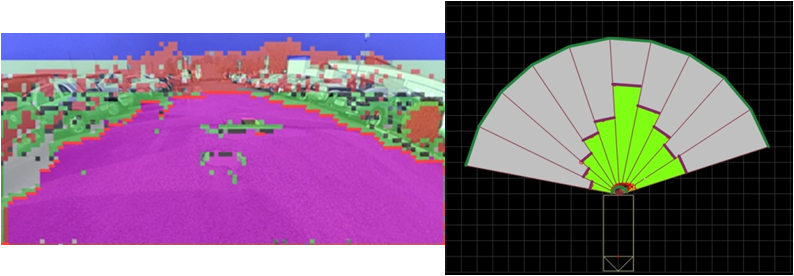}
\caption{(a) shows the road segmented from the rest of the scene use image-based segmentation, and (b) shows radial cell based freespace definition based on the road segmentation from (a)}
\label{fig:segmentation-freespace}
\end{figure}

Freespace is a feature used by most environmental sensing maps. Freespace is the area around the vehicle within the sensor’s field of view that is not occupied by objects, and is often classed as an occupancy grid map problem \cite{dollar2012pedestrian}, and is typically detected by segmenting the ground surface from other objects \cite{badino2007free}, \cite{cerri2005free}. In an occupancy-grid map approach, the freespace information is integrated and stored over time. In the case of a vector-based map representation, each object’s existence probability is updated based on the freespace measurements. Freespace is used to erase dynamic and static obstacles in the environmental sensing map which are not actively being measured or updated.  That means a good freespace model will erase dynamic obstacles from previous positions very quickly without erasing valid static information. Freespace should also erase previously detected static objects that have moved since the last valid measurement, detections that have position error in the map due to odometry update errors and false positives. Figure \ref{fig:segmentation-freespace} shows an example of image segmentation (e.g.) based freespace from a parking camera. Furthermore freespace supports a collision free trajectory search and planning especially in accumulated freespace grid maps.

Unlike other sensor types, vision systems can provide different, independent methods of estimating vehicle freespace. For example, an alternate method to determine camera-based freespace is to make use of 3D point cloud and its corresponding obstacle information. However, it also reconstructs features on the road surface around the vehicle. The features that are reconstructed that are associated with the ground can be used to provide valuable freespace information. If there is a feature reconstructed that is associated with the ground, then it is a reasonable assumption that the area between that point and the sensor (camera) is not obstructed by an object, and it can be used to define a freespace region around the host vehicle. As these methods are independent and complementary, it can also be beneficial to fuse these techniques (with themselves, and with freespace provided by other sensors, such as ultrasonics) in order to increase the accuracy and robustness of the freespace measurement.

\subsection{Other vision functions}

There are several other areas that computer vision techniques can support in the automated parking space. Visual Odometry is a task that is strongly linked to the depth estimation described in Section 4.1, via visual SLAM/bundle adjustment techniques \cite{fuentes2015visual}, although there are other methods for visual odometry \cite{scaramuzza2008appearance}. While vehicle odometry is available on vehicle networks (CAN/FlexRay), relying solely on these signals is inaccurate due to delays over the network, signal limitations and inaccuracies (e.g. relying on accelerometers), and limited degrees of freedom (typically only velocity and heading). In automated parking, the quality of the odometry is critical to user comfort and parking accuracy - as the odometry is improved, parking can be completed in fewer individual manoeuvres, and the final positions is closer to the target location.

\begin{figure}[!t]
\centering
\includegraphics[width=\columnwidth]{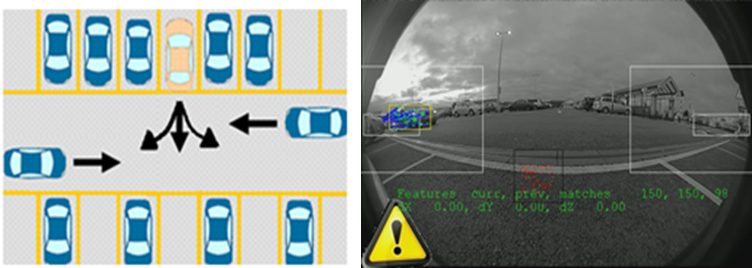}
\caption{Example of a crossing traffic situation on parking space exit, and (b) shows a screenshot of an algorithm for the detection of crossing traffic}
\label{fig:cta-illustration}
\end{figure}

Crossing traffic alert algorithms are designed to detect traffic that is potentially a threat to the host vehicle in certain critical situations at junctions, such as at a T-junction (particularly if there is limited visibility) \cite{savage2013crossing}. The need for a crossing traffic alert function on vehicles is obvious: according to the road safety website of the European Commission \cite{vacek2007road}, 40 to 60 \% of the total number of accidents occur at junctions. However, in the specific context of automated parking, crossing traffic detection is important to restrict the motion of the host vehicle, in particular when exiting a parking space, as shown in first part of Figure \ref{fig:cta-illustration}. Second picture of Figure \ref{fig:cta-illustration} shows an example of an algorithms for the detection of crossing vehicles, based on optical flow with host odometry compensation.

Other than the detection of parking slot markings discussed earlier, it is also important to be able to detect other road markings, such as arrows and disabled parking markings \cite{vacek2007road}, and road signs \cite{mogelmose2012vision}, which allows the autonomously parking vehicle to follow the rules defined in the parking area.

\subsection{A note on accuracy}

Finally for this section, we discuss briefly the accuracy required from computer vision to support the parking functions. There are two types of accuracy of the algorithms for a parking system namely detection accuracy and localisation accuracy. Computer vision benchmarks typically focus on the former but from the parking perspective, localisation is very important as well. Of course, accuracy requirements are driven by the desired functionality; conversely, the error rate of algorithms can define the feasibility of specific functions. For example, for current generation parking systems, in which the system parks in places where typically a driver could normally park, accuracy of depth of detections in the region of 10 to 15cm is adequate, as long as the standard deviation is low (~5cm). However, in the future, we may envisage automated cars parking in places dedicated to the storage of fully autonomous vehicles. In such a case, there would be a strong desire for the vehicles to park as closely as possible, thus maximising the vehicles parked per area, and as such detection accuracies of 5cm or lower would be strongly desirable.

If we discuss emergency braking in a parking scenario, the braking must only happen at the point that collision would otherwise be unavoidable, perhaps when the obstacle is only 20cm from the vehicle. The reason is quite straightforward: in a parking scenario manoeuvre is slow, and often there are many obstacles, thus allowing a larger braking distance will lead to annoying and confusing vehicle behaviour for the driver. Therefore depth accuracies of, for example, pedestrian detection should be in the region of 10cm to allow for vehicle motion and system latencies. Of course, the lower the braking distance the better, as the system can brake later while maintaining safety.  Park slot marking detection requirements are stringent as the car has to be perfectly parked, as shown in Figure \ref{fig:parking-accurate}. In such cases, localisation errors of 30cm in direction of the camera optical axis and 10cm in direction laterally of the camera position are desirable. 

These are discussion points. The actual accuracy requirement that needs to be achieved from any given algorithm must be decided upon by taking into account the exact end user functions of the parking system, the limitations in computational power of an embedded system, and the algorithmic feasibility of the function. Additionally, while the requirements for accuracy can be very high (recalling that in computer vision we natively detect angles, and not depth), within a fusion system the accuracies of other sensors, such as ultrasonics, can be used to support the camera based detection in a complementary manner.

\section{The Automated Parking System}

\subsection{Automated Parking System Considerations}

\begin{figure}[!t]
\centering
\includegraphics[width=\columnwidth]{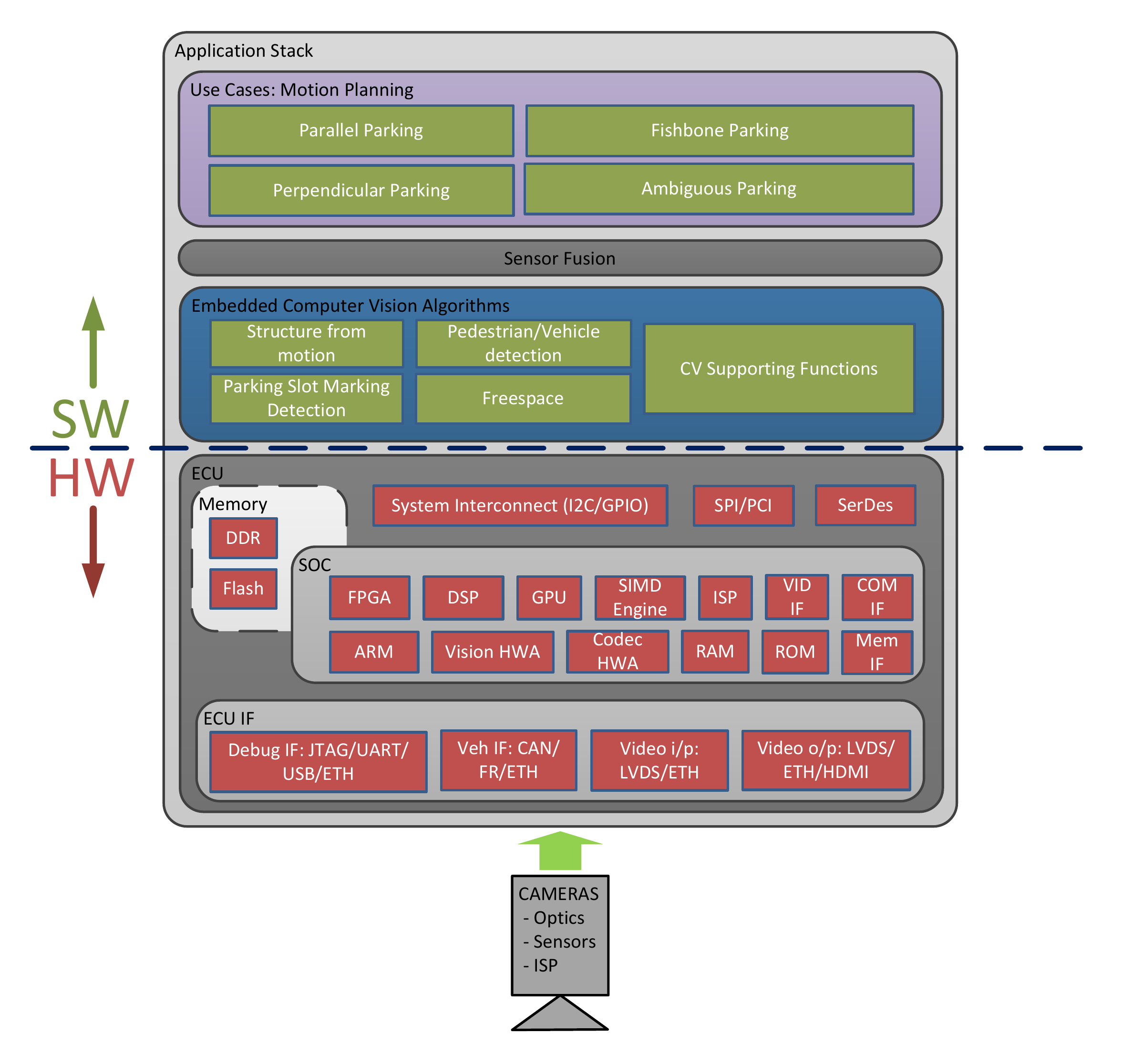}
\caption{Application Stack for full Automated Parking system using cameras.}
\label{fig:applicationStack}
\end{figure}

As discussed in section 2, 3 \& 4 as well as illustrated in Figure \ref{fig:applicationStack} there are many factors that influence the specification of a vision or partly vision (fusion) based automated parking system. Few parts of the system can be considered in isolation due to the the majority of choices having a system wide impact. A simple example is the selection of camera pixel resolution which can impact the potential use cases achievable by the system through both hardware and software; camera resolution impacts ISP selection, SerDes selection, memory bandwidth requirements, memory requirements, computational requirements of computer vision, accuracy and range performance of system, low light performance and display requirements. Therefore some limitations in terms of hardware, use cases and computer vision algorithms need to be understood and defined as illustrated in Figure \ref{fig:decisionFlow}.

From a hardware perspective the main variables, bearing in mind that thermal, power, cost are within tolerance, is the selection of the camera imager and ECU processing SOC. Typical automotive imagers for surround view applications are moving away from 1MP to 2-4MP resolutions. The challenge in increasing resolution is do so while maintaining or preferably improving low light sensitivity. This is critical for the availability and thus usability of camera based automated parking system. The extra pixel resolution improves the accuracy and range of the system allowing for increased parking use cases, more robust parking performance and increased speed availability. Once an image is formed the key is to process as many computer vision functions in parallel at as a high a frame rate and high a resolution as possible. This is where the SOC selection is critical. There are always trade-offs to reduce system load including the deployment of intelligent state machines to ensure only critical computer vision algorithms are running, downscaling and skipping of processed images to reduce loading. These trade-offs are becoming less restrictive as pixel level processing, that is generally 60-70\% of the loading of any computer vision algorithm, traditionally performed on hardware vector processing engines or DSPs are being superseded by specific computer vision hardware accelerators. These computer vision accelerators for processes such as dense optical flow, stereo disparity and convolutions are capable of much higher pixel processing throughput at lower power consumption at the expense of flexibility.

The use cases required to be covered by the system also plays a significant role in system specification. The automated parking use cases to be covered in turn define the requirements for the detection capability, accuracy, coverage, operating range, operating speed and system availability amongst others. This impacts the sensor and SOC selection but most significantly it defines the required performance from the computer vision algorithms in order to be able to achieve the functionality. For example automated perpendicular parking between lines requires many computer vision algorithms to be working in parallel, with the required accuracy and robustness to achieve a reliable and useful function. Firstly a line marking detection algorithm is required to be performing up to a speed and detection range that is practical for automated parking slot searching. In parallel an algorithm such as structure from motion is required to ensure that there is no object (parking lock, cone, rubbish bin etc.) in the slot while also measuring the end position of the slot which might be in the form of a Kerb. Pedestrian detection is also a nice addition to reduce but not remove the burden of supervision on the user during the parking manoeuvre (Level 2). These computer vision functions require support from online calibration algorithms and soiling detection functions to both operate and understand when they are not available so the system and thus user can be informed. The camera information is typically fused over time as well as with other range sensor information such as ultrasonic or radar data to improve system robustness, accuracy and availability. However, there are some detections required that only cameras can achieve such as classification. The more functions that a camera can fulfil reduces system costs as surround view cameras are becoming a standard sensor, therefore the more functions that they can achieve the less supporting sensors are required.

\subsection{A glimpse into next generation} \label{sec:future}

Computer vision has witnessed tremendous progress recently with deep learning, specifically convolutional neural networks (CNN). CNNs has enabled a large increase in accuracy of object detection leading to better perception for automated driving \cite{geiger2013vision}. It has also enabled dense pixel classification via semantic segmentation which was not feasible before \cite{cordts2016cityscapes}. Additionally there is a strong trend of CNN achieving state-of-the-art results for geometric vision algorithms like optical flow \cite{ilg2016flownet}, structure from motion \cite{ummenhofer2016demon} and re-localisation \cite{naseersemantics}. The progress in CNN has also led to the hardware manufacturers to include a custom HW IP to provide a high throughput of over 10 Tera operations per second (TOPS). Additionally the next generation hardware will have dense optical flow and stereo HW accelerators to enable generic detection of moving and static objects.

From a use case perspective, the next step for parking systems is to make them truly autonomous, which will allow a driver to leave a car to locate and park in an unmapped environment without any driver input. In addition to this, the vehicle should be able to exit the parking slot and return to the driver safely. Cameras can play a very important role in the future of automated parking systems, providing important information about the vehicle's surroundings. This includes information like object and freespace data, parking slot marking detection, pedestrian detection for fusion with other sensor technologies. 

\begin{figure}[!t]
\centering
\includegraphics[width=0.8\columnwidth]{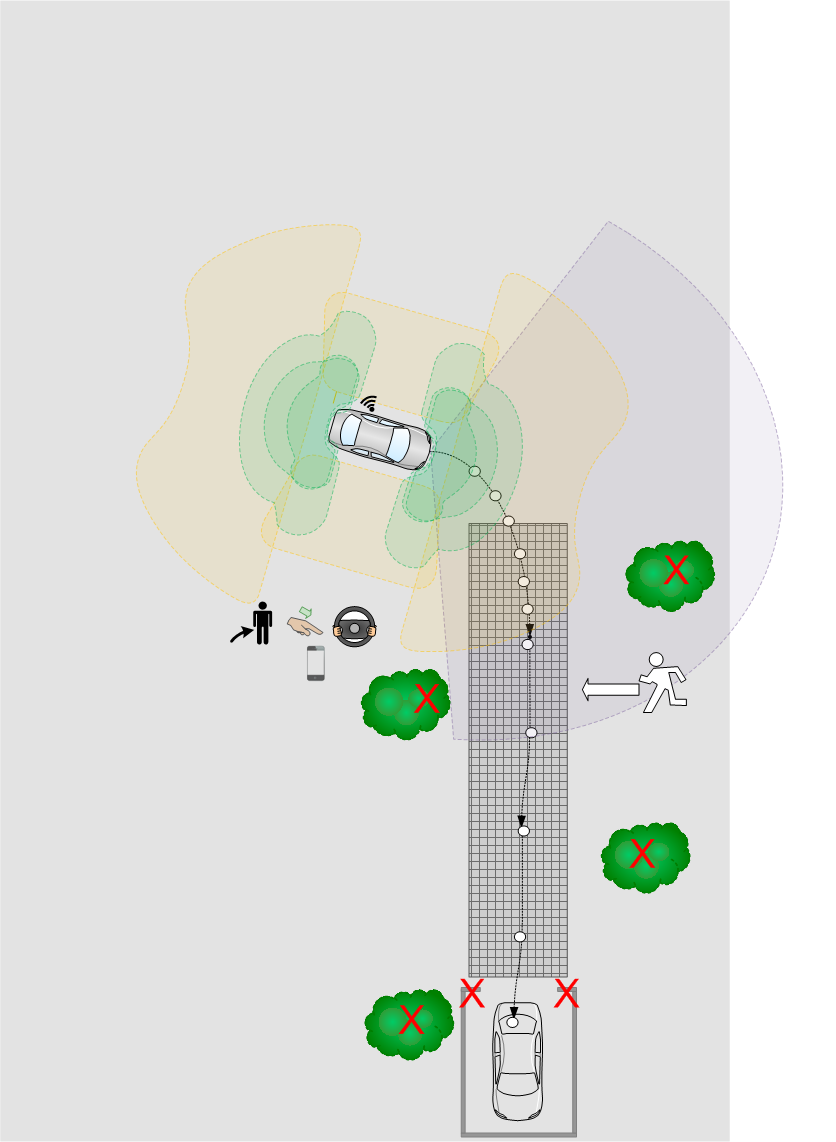}
\caption{Example of a Park4U Home Use Case where vision based systems uses landmarks to localise the car to an already stored trajectory to navigate autonomous into the home parking slot}
\label{fig:park4u}
\end{figure}

As discussed in this article current automated parking systems take control of the vehicle after recognition and selection of the parking slot by the user. The state of the system during slot search is essentially passive. The trend and challenge for the future is the automation of slot search itself in order to allow for complete vehicle parking automation including search, selection \& park all in a robust, repeatable and safe way. These automated parking scenarios can be classified as follows: 1) Automated Parking in a known area and 2) Automated Parking in an unknown area. Automated Parking in known areas typically involves the driver “training” the automated parking system with a parking trajectory (see Figure \ref{fig:park4u}). During this training the sensors locate the landmarks in the scene and record the desired trajectory driven by the driver against these landmarks. The automated parking system can recognize the scene when it returns and uses the trained information to automatically localize the vehicle to the stored trajectory allowing for automated parking.  A variant of such functionality is for example “Park4U home ”. 

Parking using a recorded trajectory poses many significant obstacles. Objects present during the training sequence may not exist during replay of the manoeuvre, for example cars, rubbish bins etc and this means that these obstacles are suitable for localisation. Even worse, objects can move very slightly in the scene between training and replay. If the system cannot identify that these objects have moved and uses them for localisation, it can result in poor trajectory replay. Objects can be moved to obstruct the trajectory during replay and this can simply result in an abort of the manoeuvre while more intelligent avoidance techniques to rejoin trajectory around the obstacle require not only extended sensor range but knowledge of drivable area and also potentially information about the private and public road divide. The complexity of traversing a trained trajectory can be increased by differences between the training and replay not only in the structure of the scene itself but also the weather and lighting conditions. These condition changes can result in vastly different views of the same scene, particularly from a visual sensor, making identification of and localisation within the “home zone” very difficult.

Automated Parking in unknown areas requires the automation of the search, selection and parking of the vehicle without the car having any prior stored trajectory. In terms of complexity this is a significant step over automated parking in known areas. Automated Parking in unknown areas was introduced by Valeo at the Frankfurt motor show via the name Valeo Valet Park4U \cite{ValeoParkingRef}.

The challenges to realize the jump to new automation levels is to extend vision based automated parking systems in terms of ego vehicle localisation (SLAM) and allow for accurate identification of stored home area. To reach the highest levels of automation for automatic parking systems it is clear that a combination of  sensor technologies (Camera, Ultrasonic, Radar or Lidar) is required to reach the maximum of accuracy, reliability in ego localisation, detection and prediction of the environment.

\section{Conclusion}

Automated driving is a rapidly growing area of technology and many high end cars have begun to ship with self-parking features. This has led to improved sensors and massive increase in computational power which can produce more robust and accurate systems. Government regulating bodies like EuroNCAP and NHTSA are introducing progressive legislation towards mandating safety systems, in spite of challenges in liability, and are starting to legislate to allow autonomous vehicles on the public road network. Camera sensors will continue to play an important role because of its low cost and the rich semantics it captures relative to other sensors. In this paper, we have focussed on the benefits of camera sensors and how it enables parking use cases. We have discussed the system implementation of an automated parking system with four fisheye cameras which provides 360 \degree view surrounding the vehicle. We covered various aspects of the system in detail including embedded system components, parking use cases which need to be handled and the vision algorithms which solve these use cases. As the focus on computer vision aspects, we have omitted the details of sensor fusion, trajectory control and motion planning. 

\section*{Acknowledgement}

All screenshots of computer vision algorithms are from products that have been developed by Valeo Vision Systems. Thanks to B Ravi Kiran (INRIA France), Prashanth Viswanath (TI), Michal Uricar (Valeo), Ian Clancy (Valeo) and Margaret Toohey (Valeo) for reviewing the paper and providing feedback.

\bibliography{mybibfile}

\end{document}